\documentclass[10pt,journal,compsoc]{IEEEtran}
\ifCLASSOPTIONcompsoc
  \usepackage[nocompress]{cite}
\else
  \usepackage{cite}
\fi

\usepackage{graphicx}
\usepackage[ruled]{algorithm2e}
\usepackage{bm}
\usepackage{multirow}
\usepackage{amsmath}
\usepackage{amsfonts}
\usepackage{amsthm}
\usepackage{xcolor,colortbl}
\usepackage{nicefrac}
\usepackage{placeins}
\usepackage[hyphens]{url}
\usepackage{enumitem}

\newcommand{\E}{\mathbb{E}}
\newcommand{\R}{\mathbb{R}}

\definecolor{red}{HTML}{f5988a}
\newcommand{\CR}{\cellcolor{red}}
\definecolor{blue}{HTML}{7eb5d6}
\newcommand{\CB}{\cellcolor{blue}}
\definecolor{green}{HTML}{b2d476}
\newcommand{\CG}{\cellcolor{green}}
\definecolor{gray}{HTML}{cfcfcf}
\newcommand{\CK}{\cellcolor{gray}}
\definecolor{yellow}{HTML}{ffd998}

\definecolor{purple}{HTML}{988ed5}
\newcommand{\CPRPL}{\cellcolor{purple}}
\definecolor{pink}{HTML}{ffb5b8}

\begin{document}

\title{Accelerating Federated Learning \\ with a Global Biased Optimiser}

\author{Jed Mills,
        Jia Hu,
        Geyong Min,
        Rui Jin,
        Siwei Zheng,
        Jin Wang
\IEEEcompsocitemizethanks{\IEEEcompsocthanksitem{ J. Mills, J. Hu, G. Min, R. Jin, S. Zheng and J. Wang are with the Department of Computer Science, University of Exeter, Exeter, UK.\protect\\
E-mail: \{jm729, j.hu, g.min, rj390, sz413, jw855\}@exeter.ac.uk.}. 
\IEEEcompsocthanksitem{Corresponding authors: Jia Hu, Geyong Min.}
\IEEEcompsocthanksitem{This paper has been accepted by IEEE Transactions on Computers.}
\IEEEcompsocthanksitem{ Source code: https://github.com/JedMills/FedGBO. }
}
}

\IEEEtitleabstractindextext{%
\begin{abstract}
Federated Learning (FL) is a recent development in distributed machine learning that collaboratively trains models without training data leaving client devices, preserving data privacy. In real-world FL, the training set is distributed over clients in a highly non-Independent and Identically Distributed (non-IID) fashion, harming model convergence speed and final performance. To address this challenge, we propose a novel, generalised approach for incorporating adaptive optimisation into FL with the Federated Global Biased Optimiser (FedGBO) algorithm. FedGBO accelerates FL by employing a set of global biased optimiser values during training, reducing `client-drift' from non-IID data whilst benefiting from adaptive optimisation. We show that in FedGBO, updates to the global model can be reformulated as centralised training using biased gradients and optimiser updates, and apply this framework to prove FedGBO’s convergence on nonconvex objectives when using the momentum-SGD (SGDm) optimiser. We also conduct extensive experiments using 4 FL benchmark datasets (CIFAR100, Sent140, FEMNIST, Shakespeare) and 3 popular optimisers (SGDm, RMSProp, Adam) to compare FedGBO against six state-of-the-art FL algorithms. The results demonstrate that FedGBO displays superior or competitive performance across the datasets whilst having low data-upload and computational costs, and provide practical insights into the trade-offs associated with different adaptive-FL algorithms and optimisers.
\end{abstract}

}

\maketitle

\IEEEdisplaynontitleabstractindextext

%
\IEEEpeerreviewmaketitle

\ifCLASSOPTIONcompsoc
\IEEEraisesectionheading{\section{Introduction}\label{sec:introduction}}
\else
\section{Introduction}
\label{sec:introduction}
\fi

\IEEEPARstart{D}{ue} to increasing public concern regarding data privacy, Federated Learning (FL) has emerged as a sub-field of machine learning that aims to collaboratively train a model in a distributed fashion, without the training data leaving the devices where it was generated. In the real world, FL deployments can range from `cross-device' scenarios where huge numbers of client devices (such as smartphones) participate, to `cross-silo' scenarios with fewer clients (such as banks and hospitals) \cite{AdvancesInFL}. As data does not leave client devices in FL, users receive a much higher degree of data-privacy compared to traditional training which would require clients to upload this sensitive data for centralised learning in the cloud.

Despite the highly attractive data-privacy benefits of FL, there exist several significant challenges to designing effective and efficient algorithms for the cross-device scenario. These challenges include:

\begin{itemize}[leftmargin=*]
\itemsep0.5em
\item{\textbf{Heterogeneous client data:} As data is generated by each client, and data shuffling is prohibited for privacy reasons, client data is highly non-Independent and Identically Distributed (non-IID). However, ML algorithms usually assume training samples are drawn IID from the true distribution, and non-IID client data has been shown to harm convergence in FL \cite{FedAvgPaper,AdaptiveFedOpt,SCAFFOLD}.}

\item{\textbf{Client networking constraints:} There are a huge number of (typically wirelessly-connected) devices such as smartphones in the cross-device scenario, and the bandwidth between the clients and server is usually low (especially considering the asymmetric bandwidth of the network edge). Furthermore, clients are unreliable and can connect to and disconnect from the FL process arbitrarily.}

\item{\textbf{Long training times:} For real-world deployments, clients may have to meet criteria in order to join the FL process. For example, in previous deployments using smartphones \cite{FLAtScale}, clients had to be connected to the internet via wifi and charging in order to participate. Therefore, the time spent waiting for clients to become available for training causes rounds of FL training to take a large amount of real-time. On top of this, client devices may have low computational power (such as embedded devices), further increasing training time.}
\end{itemize}

\begin{table*}[h]
\centering
\caption{Popular state-of-the-art FL algorithms (plus FedAvg), with per-client download and upload cost and per-client memory requirements. $\bm{x}$ denotes the FL model, and $\bm{s}$ denotes the federated optimiser values. }
\begin{tabular}{c | c c c c }
	Algorithm & Approach & Download & Upload & Memory \\
	\hline 
	FedAvg \cite{FedAvgPaper} & Averages client models & $\bm{x}$ & $\bm{x}$ & $\mathcal{O}({|\bm{x}|})$ \\
	FedProx \cite{FedProx} & Proximal term for local objectives & $\bm{x}$ & $\bm{x}$ & $\mathcal{O}(2|\bm{x}|)$ \\
	FedMAX \cite{FedMAX} & Max-entropy term for local objectives & $\bm{x}$ & $\bm{x}$ & $\mathcal{O}(|\bm{x}|)$ \\ 
 	AdaptiveFedOpt \cite{AdaptiveFedOpt} & Server-only optimiser & $\bm{x}$ & $\bm{x}$ & $\mathcal{O}(|\bm{x}|)$ \\
	MFL \cite{MFL} & Averages client models and optimisers & $\bm{x}, \bm{s}$ & $\bm{x}, \bm{s}$ & $\mathcal{O}(|\bm{x}| + |\bm{s}|)$ \\
	Mimelite \cite{Mime} & Unbiased global optimiser & $\bm{x}, \bm{s}$ & $\bm{x}, \bm{\nabla f}$ & $\mathcal{O}(2|\bm{x}| + |\bm{s}|)$ \\
	FedGBO (Ours) & Biased global optimiser & $\bm{x}, \bm{s}$ & $\bm{x}$ & $\mathcal{O}(|\bm{x}| + |\bm{s}|)$
\end{tabular}
\end{table*}

\noindent To tackle these challenges, McMahan et al. \cite{FedAvgPaper} proposed the seminal Federated Averaging (FedAvg) algorithm. In FedAvg, the server stores a single global model. The global model is downloaded by clients, who perform a given number of steps of Stochastic Gradient Descent (SGD) on the global model, using their locally-stored data. Clients then upload their new unique models to the server, which averages them together to produce the next round's global model. The authors of \cite{FedAvgPaper} showed that increasing the number of local SGD steps ($K$) increases the convergence rate of the global model, and hence the communication-efficiency of FedAvg. However, they also showed that increasing $K$ gave diminishing returns in terms of convergence rate, and that non-IID client data harms the convergence and final accuracy of the global model. Therefore, there has been much research interest in developing new FL algorithms that exhibit superior convergence (in terms of communication rounds) compared to FedAvg in the face of non-IID client data. Although most works consider training Deep Neural Networks (DNNs) in a round-based synchronous fashion, some works propose asynchronous algorithms to reduce training time \cite{SAFA,BAFL}, and for other models such as Random Forests \cite{FedForest} and Multi-Armed Bandits \cite{FedMAB}.

When clients perform $K > 1$ steps of SGD during FedAvg, each client model moves towards the minimiser of its local objective. If client data is non-IID, the models exhibit the problem of `client-drift' \cite{SCAFFOLD}: each client's model moves towards a unique, disparate minimiser. The average of these client minimisers may \textit{not} be the same as the minimiser that would be produced by centralised training on pooled client data. Client-drift harms the convergence and final performance of the federated model, and several algorithms have been proposed to address this key problem \cite{SCAFFOLD,Mime,FedDyn}. In this paper, we propose the Federated Global Biased Optimiser (FedGBO) algorithm which uses client-side adaptive optimisation to accelerate training. The optimiser values are kept constant during local training, and updated globally by the server at the end of each round using the clients' biased gradient updates. The locally-fixed  optimiser values reduce client-drift by lowering the variance of client updates, and allow a reduction in communication and computing overhead compared to competing algorithms.

The main contributions of this paper are as follows:
\begin{enumerate}[leftmargin=*]
\itemsep0.5em
\item{We propose a new FL algorithm, Federated Global Biased Optimiser (FedGBO). FedGBO is compatible with a variety of adaptive optimisers, and is the first to use locally-applied global biased adaptive parameters, which addresses the problem of client-drift whilst having low computation and communication costs.}

\item{We formulate the global-model updates of FedGBO as biased updates to a centralised model and optimiser. This formulation allows us to extend the convergence proof from previous optimisation algorithms in the centralised setting to FedGBO. To show that state-of-the-art convergence rates for centralised optimisers can be recovered (plus Fl-related error-terms), we prove the convergence of FedGBO with SGDm.}

\item{We perform an extensive set of experiments comparing FedGBO to other state-of-the-art FL algorithms on 4 benchmark FL datasets (CIFAR100, Sent140, FEMNIST, Shakespeare), using 3 popular adaptive optimisers (SGDm, RMSProp, Adam). These experiments study the benefits and drawbacks of cutting-edge algorithms in terms of convergence speed, computation and communication costs, whilst providing insights relating to the choice of adaptive-FL algorithm and optimiser for real-world FL.}
\end{enumerate}

\noindent The rest of this paper is organised as follows: Section 2 covers works proposing novel FL algorithms and theoretical analysis of client-drift; Section 3 details the design of FedGBO; Section 4 provides a convergence analysis for the global model produced by FedGBO using the momentum-SGD optimiser; Section 5 presents the results of an extensive comparison between state-of-the-art FL algorithms; and Section 6 concludes the paper.

\section{Related Work}
From a theoretical perspective, some progress has been made towards analysing the problem of client-drift in FedAvg and related algorithms with $K > 1$ local steps. The authors of \cite{ConvAccTradeOffs} exactly characterised how the objective being optimised by FedAvg is altered by increasing $K$ for quadratic objectives. In \cite{LocalFixedPoint}, the authors bounded the neighbourhood of the region around the global stationary point that FedAvg could reach for any generic fixed-point function. The problem of proving dominant speedup to the true global minimiser with $K > 1$, however, remains open for all but quadratic objectives \cite{IsLocalSGDBetter}. Some works such as \cite{AdaptFLResourceConstraint} have also demonstrated improvement to global model convergence when using variable number of local steps or batch size.

One approach for addressing the problem of client-drift is the use of control variates on clients. The authors of \cite{SCAFFOLD} proposed the SCAFFOLD algorithm, which uses local and global control variates for client-based Stochastic Variance-Reduced Gradients (SVRG) \cite{SGVR}. FedDyn \cite{FedDyn} saves on the per-round communication cost of SCAFFOLD by incorporating the global control variate into the model update. However, SCAFFOLD and FedDyn require stateful clients (i.e., they must permanently store local variates), so are not compatible with the cross-device FL scenario. Later, \cite{Mime} proposed the Mime framework that also tackles the problem of client-drift. In Mime, a set of global optimiser statistics are tracked and a control variate is used to reduce client-drift and improve convergence speed. These authors proved the theoretical advantages of Mime for convex and nonconvex objectives, and demonstrated its real-world performance on benchmark FL datasets. However, Mime is limited by significantly increased communication and computational costs compared to FedAvg. 

\begin{table*}[h]
\centering
\caption{Update, Tracking and Inverse steps for three popular optimisers, with: decay rates $\beta, \beta_1, \beta_2$; (stochastic) gradients $\bm{g}$; stability constant $\epsilon$; learning rate $\eta$ and total local steps $K$; and previous and current global models $\bm{x}_t$ and $\bm{x}_{t+1}$.}
\begin{tabular}{ c | c c c c }
	Algorithm & Statistics & Update ($\mathcal{U}$) & Tracking ($\mathcal{T}$) & Inverse ($\mathcal{I}$) \\ 
	\hline
	RMSProp	& $\bm{v}$ & $\frac{\eta \bm{g}}{\sqrt{\bm{v}} + \epsilon}$ & $\bm{v} \gets \beta \bm{v} + (1 - \beta) \bm{g}$ & $\frac{1}{\eta K}(\bm{x}_t - \bm{x}_{t+1})(\sqrt{\bm{v}} + \epsilon)$ \\
	SGDm 	& $\bm{m}$ & $\eta(\beta \bm{m} + (1 - \beta) \bm{g})$ & $\bm{m} \gets \beta \bm{m} + (1 - \beta) \bm{g}$ & $\frac{1}{1 - \beta}(\frac{\bm{x}_t - \bm{x}_{t+1}}{\eta K} - \beta \bm{m})$ \\
	\multirow{2}{*}{Adam} & \multirow{2}{*}{$\bm{m}, \bm{v}$} & \multirow{2}{*}{$\frac{\eta(\beta_1 \bm{m} + (1 - \beta_1) \bm{g})}{\sqrt{v} + \epsilon}$} & $\bm{m} \gets \beta_1 \bm{m} + (1 - \beta_1) \bm{g}$ & \multirow{2}{*}{$\frac{1}{1 - \beta_1} \left( \frac{(\bm{x}_t - \bm{x}_{t+1})(\sqrt{\bm{v}} + \epsilon)}{\eta K} - \beta_1 \bm{m} \right)$} \\
	& & & $\bm{v} \gets \beta_2 \bm{v} + (1 - \beta_2) \bm{g}$ & \\
\end{tabular}
\end{table*}

Other authors have also investigated ways of incorporating adaptive optimisation into FedAvg to improve convergence. \cite{FLKeywordSpotting,AdaptiveFedOpt} proposed adding adaptive optimisation to the server-level update of the global model. This algorithm (henceforth referred to as AdaptiveFedOpt) can converge faster than FedAvg in some settings and has local computation and communication costs as low as FedAvg, but does not benefit from client-side adaptive optimisation. Alternatively, \cite{MFL} proposed Momentum Federated Learning (MFL) that uses momentum-SGD (SGDm) on clients, and also averages the momentum parameters each round. MFL provides significant speedup compared to FedAvg, but naturally increases communication and local computation costs. Similarly, \cite{CommEffFLIoT} averaged client Adam parameters alongside model weights to improve convergence speed, with added compression. For the rest of this paper, we refer to any algorithm that averages client optimiser parameters alongside their model parameters as MFL.

Another approach for improving FedAvg's convergence is to modify the client objective functions. \cite{FedProx} proposed FedProx, which adds a proximal term to discourage the local model from straying too far from the global model. Alternatively, \cite{FedMAX} added an error term maximising the entropy of the penultimate-layer activations of DNNs in their FedMAX algorithm. Doing so makes the model activations between non-IID clients more similar during training, thus improving model aggregation. \cite{ClientSideOptStrats} provides a recent survey and comparison of state-of-the-art client-side optimisation algorithms within FL. 
\color{black}

\section{The FedGBO Algorithm}
\subsection{Algorithm Design}
The Federated Global Biased Optimiser (FedGBO) algorithm operates in rounds similar to FedAvg. As presented in Algorithm 1, in each communication round of FedGBO, a random subset of clients are selected to participate. This selection forces partial client participation in the experimental setting, however in a real-world deployment, clients devices will participate if they are available for training (e.g., for smartphone clients this may be when they are charging and connected to WiFi) \cite{FLAtScale}. FedGBO is a generic FL algorithm compatible with any adaptive optimiser that can be represented as an update ($\mathcal{U}$), tracking ($\mathcal{T}$), and inverse ($\mathcal{I}$) step. Table 2 presents $\mathcal{U}, \mathcal{T}, \mathcal{I}$ for three popular optimisers: RMSProp \cite{RMSProp}, SGDm, and Adam \cite{Adam}.

\begin{algorithm}[h]
\caption{FedGBO}
\textbf{input:} global model $\bm{x}_0$, optimiser $\bm{s}_0$, learning rate $\eta$ \\
\For{\rm{round} $t = 1 \ \textbf{to} \ T$}{
	select clients $\mathcal{S}_t$ \\
	\For{\rm{client} $i \in \mathcal{S}_t$ \rm{in parallel}}{
		download global model $\bm{x}_t$, and optimiser $\bm{s}_t$	\\
		initialise local model, $\bm{y}^i_0 \gets \bm{x}_t$  \\
		\For{\rm{local SGD step} $k = 1 \ \textbf{to} \ K$}{
			compute minibatch gradient $\bm{g}^i_t$ \\
			$\bm{y}_k^i \gets \bm{y}_{k-1}^i - \mathcal{U}(\eta, \bm{g}^i_t, \bm{s}_t)$
		}
		upload local model $\bm{y}^i_K$ to server
	}
	update global model $\bm{x}_{t+1} \gets \frac{1}{|S_t|} \sum_{i \in S_t} \bm{y}^i_K$ \\
	compute biased gradient $\tilde{\bm{g}}_t \gets \mathcal{I}(\bm{x}_t, \bm{x}_{t+1}, \bm{s}_t, K, \eta)$\\
	update global optimiser values $\bm{s}_{t+1} \gets \mathcal{T}(\tilde{\bm{g}}_t, \bm{s}_t)$
}
\end{algorithm}

Participating clients download the global model $\bm{x}_t$ and optimiser values $\bm{s}_t$, and then perform $K$ steps of adaptive optimisation using the generic update step $\mathcal{U}$ and locally-computed stochastic gradients. After local training, all participating clients upload their local models to the server. The server computes the new global model $\bm{x}_{t+1}$ as an average of client models. It performs the generic inverse-step $\mathcal{I}$ to compute the average gradient during the client local updates. The server then updates the global statistics using the generic tracking step $\mathcal{T}$. 

An alternative implementation of FedGBO (that would not require optimisers to be represented using an inverse step $\mathcal{I}$) would update the local model $\bm{y}^i_k$ on clients, whilst also maintaining a copy of the average local gradient (which would be the same size as the model used, $|\bm{x}|$). This average local gradient would then be uploaded to the server, and could be used to update the global model and global statistics instead of performing an inverse step $\mathcal{I}$. However, FedGBO is designed to be lightweight in terms of computation and memory cost for clients, as real-world FL scenarios can involve low-powered clients such as IoT devices. Therefore, our design decreases the memory and total computational cost for clients as the average gradient does not have to be stored alongside the local model $\bm{y}^i_k$.

\begin{figure*}[h]
\includegraphics[width=\textwidth]{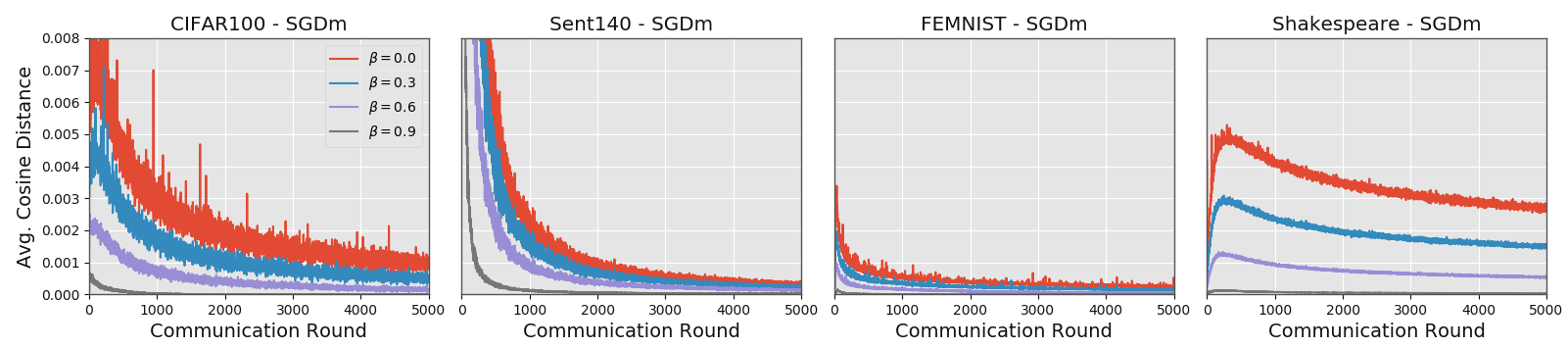}
\caption{Average cosine distances between client models uploaded at the end of each communication round. Curves are averages over 5 random trials. Note FedGBO with $\beta = 0$ is equivalent to FedAvg. }
\end{figure*}

FedGBO shares some design similarities with other adaptive-FL algorithms. We describe the design differences between FedGBO and these algorithms below.
\begin{itemize}[leftmargin=*]
\item{\textit{AdaptiveFedOpt:} clients download a global model each round and perform vanilla SGD locally. The server generates a `psuedogradient' each round from client model uploads. This psuedogradient is fed into an adaptive optimiser that exists only on the server to update the global model at the end of each round.}

\item{\textit{MFL:} clients download a global optimiser each round (like FedGBO). However, the optimiser values are allowed to change during the client loop (as opposed to fixed). The unique client optimisers are uploaded to the server each round along with the client models. The server averages the client optimisers to make the next round's global optimiser (as opposed to updating it via the $\mathcal{I}$ and $\mathcal{T}$ steps) alongside averaging the client models to produce the next global model.}

\item{\textit{Mimelite:} clients download a global model and optimiser each round and keep the values fixed during the client loop (like in FedGBO). However, clients also compute a full-batch gradient ($\bm{\nabla} f$) before local training and send this gradient along with the model to the server after local training (FedGBO does not compute or send $\bm{\nabla}f$). The server uses $\bm{\nabla}f$ to update the global optimiser (as opposed to updating it via the $\mathcal{I}$ and $\mathcal{T}$ steps)}.
\end{itemize}

\noindent Table 1 also summaries the practical differences between the algorithms. The beneficial aspects of FedGBO compared to the other algorithms in Table 1 are: compared to the vanilla SGD used by FedAvg, FedProx, FedMAX and AdaptiveFedOpt, FedGBO's client-side optimisation can substantially accelerate model convergence; compared to MFL, FedGBO accounts for client-drift with fixed optimiser values, which can help accelerate convergence and has lower upload cost depending on the optimiser used ($50\%$ less for SGDm/RMSProp, $66\%$ less for Adam); compared to Mimelite, FedGBO does not require the expensive computation of $\bm{\nabla}f$ and has $50\%$ reduced upload cost. Reducing upload costs in FL is of particular importance due to the asymmetric bandwidth of devices at the network edge.
\color{black}

\subsection{Reducing Client Drift}
FedGBO reduces client-drift by keeping a set of global statistics (downloaded at the start of each communication round) that are not updated in the local-training loop, as shown in Algorithm 1. For FedGBO with a momentum-based optimiser, when the `decay' parameter (e.g., $\beta$ in SGDm) is large, there is less influence from client gradients in the local updates, and more influence from the global biased statistics, so the models uploaded by clients will be more similar. For adaptive-learning rate methods (e.g., RMSProp), using the same fixed adaptive parameters on all clients scales the local updates performed by clients for a given model parameter by the same value, as opposed to letting the adaptive parameters change during the local update. 

In Fig. 1 we plot the average cosine distance between client models at the end of each communication round, for four benchmark FL datasets (further details about datasets and models are given in Section 5), using FedGBO with the SGDm optimiser. When $\beta = 0$ (red curves), there is no influence from the global biased momentum, and FedGBO is equivalent to FedAvg. As explained above, Fig. 1 (a) - (d) show that increasing the decay parameter ($\beta$) causes client updates to be more aligned, due to more influence from the fixed global momentum, leading to more similar client models. Fig. 1 therefore demonstrates how FedGBO can tackle the problem of client-drift.

\section{Convergence Analysis}
In this section, we show that the updates to the global model in FedGBO with a generic optimiser can be formulated as updates to the same optimiser in the centralised setting, with a perturbed gradient and statistics update. We then show that this formulation can be used to extend existing convergence analyses by applying a recent analysis of SGDm \cite{SimpleConvergeAdam} to FedGBO, recovering the same tight dependence on $\beta$ as in the centralised analysis, with added client-drift terms associated with FL.

In FL, we wish to train a model $\bm{x} \in \mathbb{R}^d$ that minimises the following objective function:
\begin{equation}
F(\bm{x}) = \underset{i}{\E} \left[ f_i(\bm{x}) = \frac{1}{n_i} \sum_{j = 1}^{n_i} f (\bm{x}; \xi_{i, j}) \right],
\end{equation}
where $f_i$, $n_i$ and $\{ \xi_{i,1}, \cdots, \xi_{i, n_i} \}$ denote the average loss, total number of samples, and training samples on client $i$, respectively. $f$ is the loss function that clients use to train $\bm{x}$. Intuitively, we wish to minimise the expected loss over all samples and all clients, as would be minimised by training in the centralised setting over pooled data. We present the individual client losses $f_i$ as a sum and the global loss $F$ as an expectation to emphasise that in FL there are a very large number of clients, each possessing a small number of local samples.

Clients are assumed to have heterogeneous local data distributions. Therefore, the local minimiser $f^*$ for any two clients $i$ and $j$ are not necessarily the same: $f_i^* \neq f_j^*$. This is the source of client drift for $K > 1$ local updates. We use the following standard assumptions to make the analysis of FedGBO tractable.
\\

\noindent \textbf{Assumption 1} (Lower bound). \emph{$F$ is bounded below by $F^*$: $F(\bm{x}) > F^*, \forall \bm{x} \in \R^d$.}
\\

\noindent \textbf{Assumption 2} (Inter-client variance). \emph{The variance of client gradients is bounded: $ \underset{i}{\E} \left[ \| \nabla f_i(\bm{x}) - \nabla F(\bm{x}) \| ^2 \right] \leq G^2, \forall \ i$.}
\\

\noindent \textbf{Assumption 3} (Gradient magnitude). \emph{The magnitude of client gradients is bounded: $\| f_i(\bm{x}; \xi) \|^2 \leq R^2, \forall \ i$.}
\\

\noindent \textbf{Assumption 4} (Lipschitz gradients). \emph{Client loss functions are $L$-smooth: $\| \nabla f_i(\bm{x}; \xi) - \nabla f_i(\bm{y}; \xi) \| \leq L \| \bm{x} - \bm{y} \|, \forall \ i$. $F$ is a convex combination of $f_i$, so $F$ is therefore also $L$-smooth.}
\\

\noindent \textbf{Assumption 5} (Intra-client variance). \emph{Client stochastic gradients are unbiased estimates of the local gradients, $\underset{\xi}{\E} \left[ \nabla f_i(\bm{x}; \xi) \right] = \nabla f_i(\bm{x}), \forall \ i$, with bounded variance: $\underset{\xi}{\E} \left[ \| \nabla f_i(\bm{x}; \xi) - \nabla f_i(\bm{x}) \|^2 \right] \leq \sigma^2, \forall \ i$.}
\\

\noindent Previously, \cite{Mime} showed that FedAvg with $K > 1$ local updates could be reformulated as centralised optimisation with a perturbed gradient, which they then use to prove the convergence of their Mime algorithm. The perturbation $\bm{e}_t$ of the centralised-training gradient $\bm{g}_t$ is defined as:
\begin{equation}
\bm{e}_t = \frac{1}{K |S|} \sum_{i = 1}^{K} \sum_{i \in S} \left( \nabla f_i(\bm{y}_{i, k-1}; \xi_{i, k}) - \nabla f_i (\bm{x}; \xi_{i, k}) \right),
\end{equation}
\noindent for local iterations $\{ y_{i,0}, \cdots, y_{i, K-1} \}$, and set of sampled clients $S$. Thus, the gradient-perturbation $\bm{e}_t$ is the average difference between client gradients (over the $K$ local steps and $S$ clients) and the gradients that would have been computed using the global model. Using this definition, we can rewrite the updates of FedGBO as perturbed updates to a centralised optimiser:
\begin{align*}
\bm{x}_t &\gets \bm{x}_{t - 1} - \tilde{\eta} \mathcal{U}( \bm{g}_t + \bm{e}_t, \bm{s}_{t - 1}), \\
\bm{s}_t &\gets \mathcal{V}(\bm{g}_{t} + \bm{e}_t, \bm{s}_{t - 1}),
\end{align*}
for generic model-update step $\mathcal{U}$, optimiser tracking step $\mathcal{T}$, and `psuedo-learning-rate' $\tilde{\eta} = K\eta$. The local updates of FedGBO use fixed global statistics in a similar manner to the fixed statistics used in Mime (albeit we use biased global statistics). As we make assumptions at least as strong as those from \cite{Mime}, we can directly use their result to bound the norm of the perturbation $\bm{e}_t$:
\begin{align}
\underset{t}{\E} \left[ \| \bm{e}_t \|^2 \right] \leq (B^2 L^2 \tilde{\eta}^2) \left( \underset{t}{\E} \left[ \| \bm{g}_t \|^2 \right] + G^2 + \frac{\sigma^2}{K} \right),
\end{align}
where $B \geq 0$ is a bound on the Lipschitzness of the optimiser $\mathcal{U}$'s update: $\| \mathcal{U}(\bm{g}) \| \leq B \| \bm{g} \|$.

Using the perturbed gradient and momentum generalisation presented above, updates of FedGBO using the SGDm optimiser can be written as:
\begin{align}
\bm{m}_t &\gets \beta \bm{m}_{t - 1} + (\bm{g}_{t - 1} + \bm{e}_{t - 1}) \\
\bm{x}_{t} &\gets \bm{x}_{t - 1} - \tilde{\eta} \bm{m}_t.
\end{align}
While the specific SGDm implementation we use in the later experiments multiplies the second term of (4) by a $(1 - \beta)$ term (we find hyperparameter choice with this implementation to be more intuitive), we drop the $(1 - \beta)$ term in (4) to simplify the proof, and assume the $(1 - \beta)$ and $\beta$ terms can be incorporated into $\tilde{\eta}$. As is typical for nonconvex analysis, we now bound the expected squared norm of the model gradient at communication round $t$. For total iterations $T$, we define $t$ as a random index with value $\{0,\cdots,T-1\}$, with probability distribution $\mathbb{P}[t = j] \propto 1 - \beta^{T - j}$. Full proofs are given in Appendix A.
\\

\begin{table*}[h]
\normalsize
\centering
\caption{Details of datasets and models used in the experiment tasks presented in Section 4 (CNN = Convolutional Neural Network, GRU = Gated Recurrent Network).}
\begin{tabular}{ c | c c c c c c }
 	\multirow{2}{*}{Task} & \multirow{2}{*}{Type} & \multirow{2}{*}{Classes} & \multirow{2}{*}{Model} & \multirow{2}{*}{Total Clients} & Sampled Clients & Mean Samples \\
	& & & & & per Round & per Client \\ 
	\hline
	CIFAR100 & Image classification & 100 & CNN & 500 & 5 & 100 \\
	Sent140 & Sentiment analysis & 2 & Linear & 21876 & 22 & 15 \\
	FEMNIST & Image classification & 62 & CNN & 3000 & 30 & 170 \\
	Shakespeare & Next-character prediction & 79 & GRU & 660 & 7 & 5573 \\
\end{tabular}
\end{table*}

\noindent \textbf{Theorem 1} (Convergence of FedGBO using SGDm) \textit{Let the assumptions A1-A5 hold, the total number of iterations $T > (1 - \beta)^{-1}$, $\tilde{\eta} = K\eta > 0, 0 \leq \beta < 1$, and using the update steps given by (4) and (5), we have:}
\begin{multline}
\E[\|\nabla F(\bm{x}_t)\|^2] \leq \frac{2(1 - \beta)}{\tilde{\eta} \tilde{T}}(F(\bm{x}_0) - F^*) + \frac{T \tilde{\eta}^2 Y}{\tilde{T}} \\
								+ \frac{2T\tilde{\eta}L(1+\beta)(R^2 + \tilde{\eta}^2 Y)}{\tilde{T}(1 - \beta)^2},
\end{multline}
where $\tilde{T} = T - \frac{\beta}{1 - \beta}$, and $Y = 18 B^2 L^2 \left( R^2 + G^2 + \frac{\sigma^2}{K} \right)$. 
\\

\noindent Theorem 1 therefore shows that the perturbed gradient and momentum framework presented above can be used to apply an existing convergence proof from an centralised optimiser to FedGBO. Next, we simplify Theorem 1 to achieve a more explicit convergence rate.
\\

\noindent \textbf{Corollary 1} (Theorem 1 simplification using iteration assumption) \textit{Making the same assumptions as in Theorem 1, and also that $T \gg (1 - \beta)^{-1}$, hence $\tilde{T} \approx T$, and using $\tilde{\eta} = (1 - \beta)\frac{2C}{\sqrt{T}}$, for $C > 0$, then}
\begin{multline}
\E[\|\nabla F(\bm{x}_t)\|^2] \leq \frac{F(\bm{x}_0) - F^*}{C \sqrt{T}} \\
+ \frac{4CL(1+\beta)(R^2 + \nicefrac{Z}{T})}{(1-\beta)\sqrt{T}} + \frac{Z}{T},
\end{multline}
where $Z = 4 C^2 (1 - \beta)^2 Y$.
\\

\noindent Corollary 1 shows that the gradients of FedGBO can be bounded for nonconvex objectives. To the best of our knowledge, this is the first convergence analysis of an FL-algorithm using SGDm for nonconvex objectives: \cite{MFL} analyses strongly-convex objectives with deterministic gradients, and \cite{AdaptiveFedOpt,Mime} analyse Adam with $\beta_1 = 0$ (i.e., RMSProp). The lack of directly-related analyses makes comparison of this analysis to existing works less clear, but Corollary 1 still provides useful insights into FedGBO's convergence.

\begin{itemize}[leftmargin=*]
\item{\textbf{Relation to centralised rate:} Comparing (7) to Theorem B.1 of \cite{SimpleConvergeAdam} shows that the $\mathcal{O}(\nicefrac{1}{\sqrt{T}})$ convergence rate can be retained (with the same state-of-the-art $(1 - \beta)$ denominator). However, (7) contains added error terms due to client-drift: $Z$ in the second term arises from the biased momentum used in FedGBO, and the third term arises from biased client gradients. }

\item{\textbf{Setting $\bm{\tilde{\eta}}$:} Scaling $\tilde{\eta}$ with $\mathcal{O}(\nicefrac{1}{\sqrt{T}})$  fortunately decreases the error due to biased client gradients and momentum with $\mathcal{O}(\nicefrac{1}{T})$ and $\mathcal{O}(\nicefrac{1}{\sqrt{T}})$, respectively. As $\tilde{\eta} = K \eta$, this indicates that either $K$ or $\eta$ could be decayed during training to balance fast convergence and final error.}

\item{\textbf{Effect of $\bm{K}$:} According to (7), there is no benefit for FedGBO's iteration complexity with $K > 1$ local steps, which is a common within FL analysis. \cite{MFL} proves the convergence of MFL by bounding the distance to centralised momentum-SGD, and this distance naturally converges to $0$ as $K \to 1$. Similarly, \cite{SCAFFOLD,Mime} improve convergence rates for convex objectives when $K > 1$ by adding Variance-Reduction (VR) to the client objective (which increases communication and computation costs). Without VR, the convergence of their algorithms do not dominate distributed SGD (with $K = 1$). However, communication complexity in FL is usually of greater importance due to slow communication. Section 5 shows that large $K$ can significantly improve the convergence rate in terms of communication rounds.. }
\end{itemize}
\color{black}

\begin{figure*}[h]
\includegraphics[width=\textwidth]{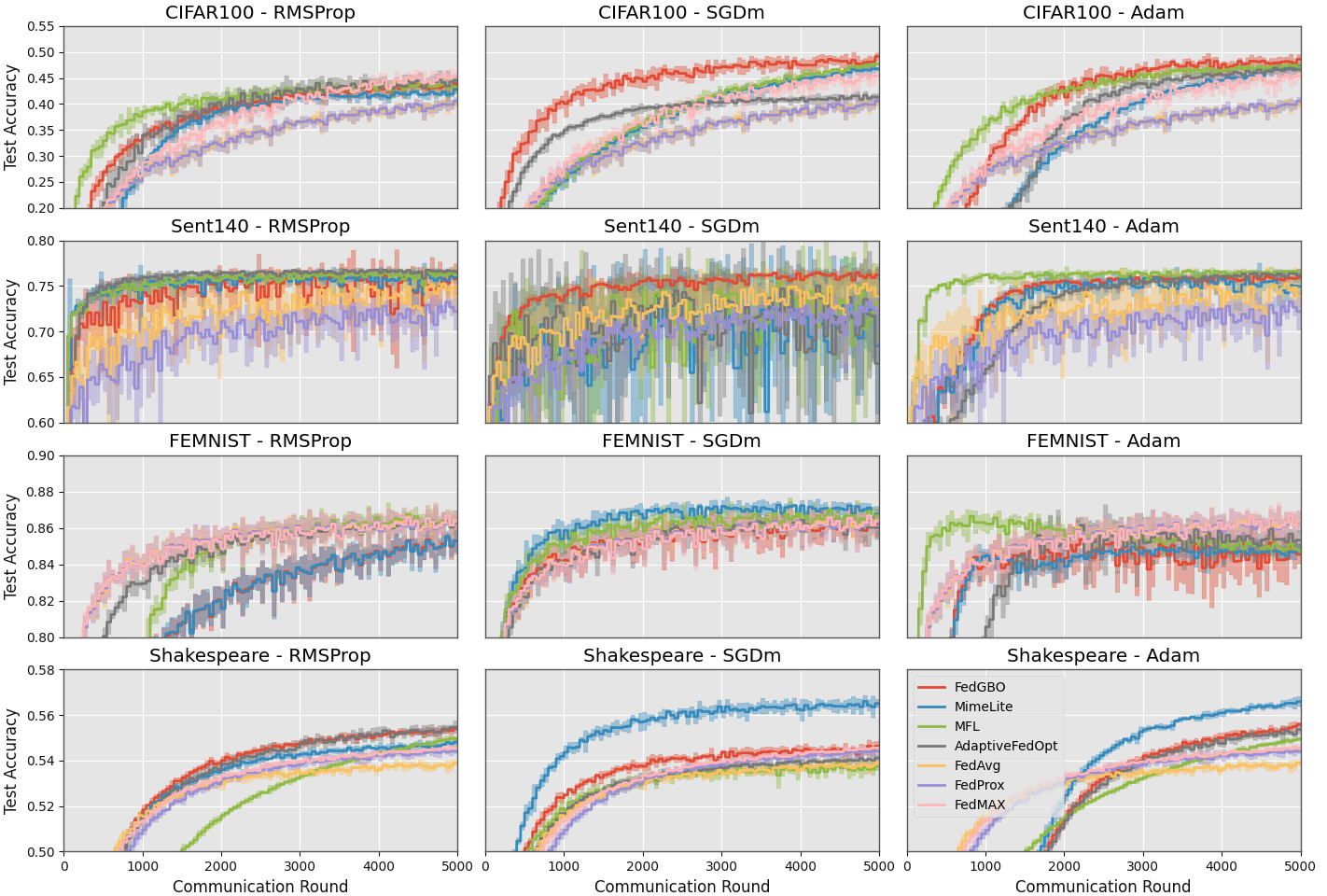}
\caption{
Comparison of adaptive FL algorithms on different FL datasets, using $K = 10$ local update steps. Lines show average over 5 random trials, shaded regions show $95\%$ confidence intervals of the mean.
\color{black}
}
\end{figure*}

\section{Experiments}
In this section, we conduct a comprehensive comparison between FedGBO and various state-of-the-art adaptive-FL algorithms (AdaptiveFedOpt \cite{AdaptiveFedOpt}, MFL \cite{MFL}, Mimelite \cite{Mime}), and non-adaptive algorithms (FedAvg \cite{FedAvgPaper}, FedProx \cite{FedProx}, FedMAX \cite{FedMAX}). The experiments are conducted on 4 benchmark FL datasets, from the domains of image classification, sentiment analysis, and language modelling. Each adaptive-FL algorithm is compatible with a variety of optimisers, so we test three of the most popular and ubiquitous: RMSProp \cite{RMSProp}, SGDm, and Adam \cite{Adam}. Although from an algorithmic perspective RMSProp and SGDm are simply special cases of Adam (with $\beta_1 = 0$ and $\beta_2 = 0$, respectively), for FL there is the important practical difference of not requiring the zeroed parameters to be communicated. All of the models and algorithms tested were implemented with Pytorch 1.7.0, and run on workstations equipped with Intel i9 CPUs and Nvidia RTX 3090 GPUs, running Ubuntu 20.04. Code for the experiments can be found in our online repository. Table 3 gives a brief overview of the datasets and models used, with more thorough details below.
\\

\noindent \textbf{CIFAR100:} a federated version of the CIFAR100 dataset consisting of $(32 \times 32)$ pixel RGB images from 100 classes. Training samples are partitioned according to the class labels into 500 workers using the Pachinko Allocation Method first used in \cite{AdaptiveFedOpt}. Like in similar FL works \cite{AdaptiveFedOpt,Mime}, we apply preprocessing to the training samples comprising a random horizontal flip ($p = 0.5$) followed by a random crop of the $(28 \times 28)$ pixel sub-image. We train a Convolutional Neural Network (CNN) with the following architecture: a $(3 \times 3 \times 32)$ ReLU Convolution layer, $(2 \times 2)$ max pooling, $(3 \times 3 \times 64)$ ReLU Convolution layer, $(2 \times 2)$ max pooling, a 512-unit ReLU fully connected layer, and softmax output, trained with batch size of 32.
\\

\noindent \textbf{Sent140:} a sentiment analysis task using Twitter posts, preprocessed using the LEAF FL benchmark suite \cite{LEAF}. Tweets are grouped by user, users with $< 10$ posts are discarded, and $20 \%$ of each users' samples are taken for the test set. We then create a normalised bag-of-words vector of length 5k (representing presence of the 5k most common token in the dataset), with each target being a vector of length two (positive or negative sentiment). Samples without any of the top 5k token are discarded. We train a linear model using batch size 8 on this dataset. Note that we do not test FedMAX on Sent140 as FedMAX is only compatible with DNNs.
\\

\noindent \textbf{FEMNIST:} a federated version of the EMNIST dataset containing $(28 \times 28)$ pixel greyscale images from 62 classes, preprocessed using LEAF \cite{LEAF}. Samples are grouped into users by the writer of the symbol, and $20\%$ of each user's samples are grouped to make a test-set. We select 3000 clients (from the maximum 3550) for use in all experiments. We train a CNN with the same architecture as the CIFAR100 model, albeit with 62 outputs, and batch size of 32.
\\

\noindent \textbf{Shakespeare:} a next-character prediction task of the complete plays of William Shakespeare, preprocessed using LEAF \cite{LEAF}. The lines from all plays are grouped by speaker (e.g., Macbeth, Lady Macbeth etc.). Speakers with $< 2$ lines are discarded, leaving 660 total clients. Lines are processed into sequences of length 80, with a vocabulary size of 79, and the last $20\%$ of each client's lines are taken to produce the test-set. We train a Gated Recurrent Unit (GRU) model on this dataset comprising a trained embedding layer with 8 outputs, two stacked GRU layers with 128 outputs each, and a softmax output layer, using a batch size of 32.
\\

\begin{figure*}[h]
\includegraphics[width=\textwidth]{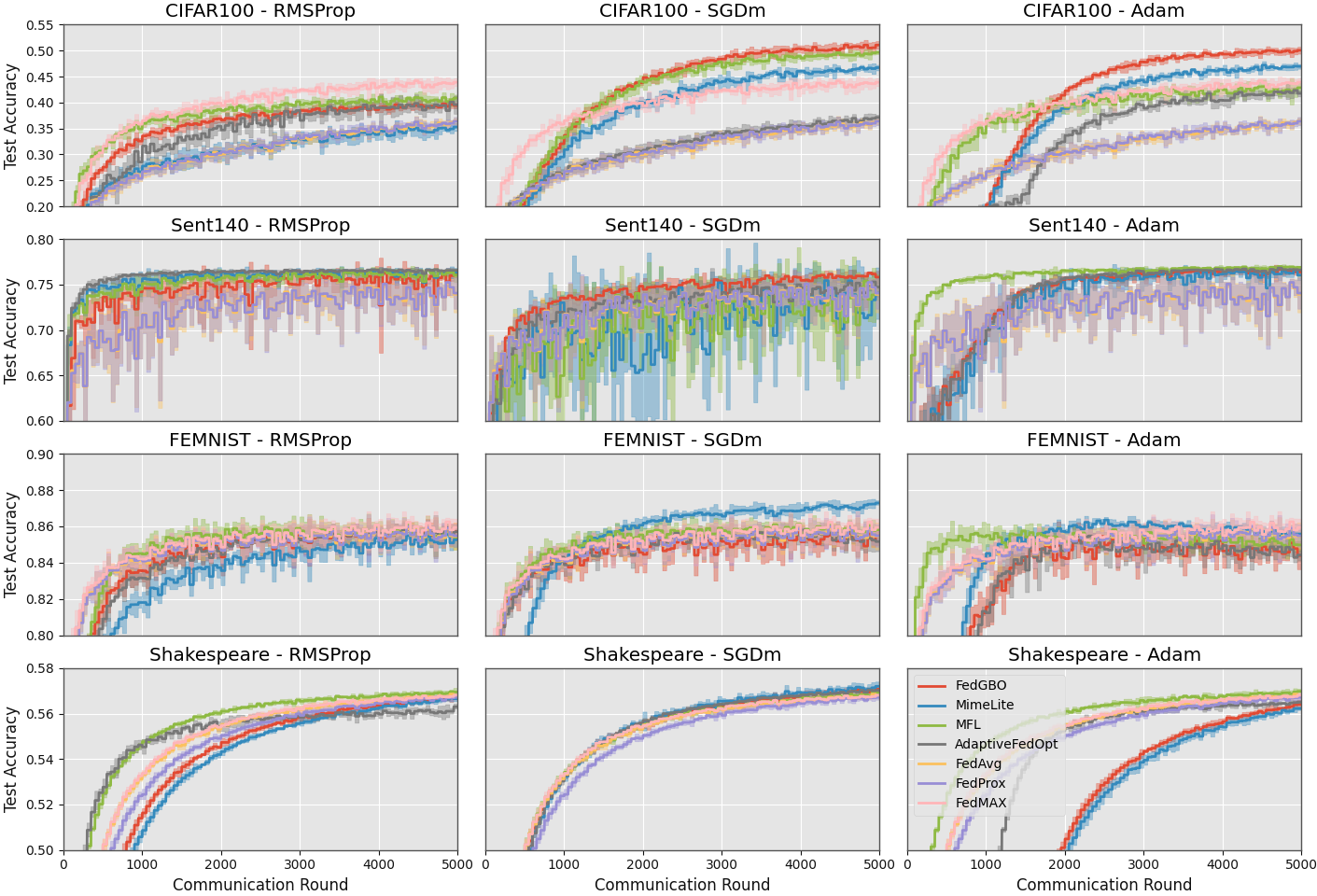}
\caption{
Comparison of adaptive-FL algorithms on different FL datasets, using $K = 50$ local update steps. Lines show average over 5 random trials, shaded regions show $95\%$ confidence intervals of the mean.
\color{black}
}
\end{figure*}

\subsection{Convergence Speed}
We first compare the convergence speed of the different adaptive-FL algorithms using each optimiser, on each dataset. For each [dataset, FL algorithm, optimiser] combination, we tuned hyperparameters via grid search in order to achieve the highest test-set accuracy within 5,000 communication rounds (evaluated every 50 rounds). We tested learning rates in the range $\eta \in [10^{1}, 10^{-5}]$, with step size of $0.1$, and optimiser parameter $\beta = \{0.3, 0.6, 0.9, 0.99, 0.999, 0.9999 \}$. For Adam, in order to keep the number of experiments performed feasible, we fixed $\beta_2 = 0.99$ for all experiments and tuned $\beta_1$ as above. For RMSProp and Adam, as has been noted previously \cite{AdaptiveFedOpt,Mime} we found that a large adaptivity parameter $\epsilon = 10^{-3}$ was required for stable convergence. Tuning the AdaptiveFedOpt algorithm took significantly more simulations compared to the other algorithms, due to the extra hyperparameter (the server learning rate, on top of client learning rate and optimiser parameter), which represents a drawback for real-world use. 

For FedAvg we tune only $\eta$. For FedProx we tune $\eta$ and proximal term $\mu$. For FedMAX we tune $\eta$ and entropy-loss-weighting $\alpha$. These algorithms do not use adaptive optimisation, so we present the same curve for each algorithm in each row of Figs. 2 and 3. The best hyperparameters for all scenarios can be found on our online code repository. In total we performed well over 1000 simulations.

In Fig. 2, we present the test-set accuracies during training for these tuned parameters using $K = 10$ local steps. We consider this a `low-$K$' regime, representing scenarios where clients can communicate more frequently with the server. Fig. 3 shows the accuracy curves for a `high-$K$' ($K = 50$) regime. We now discuss the practical insights provided by these experiments.
\\

\noindent \textbf{Choice of FL algorithm:} Comparing the rows of Figs. 2 and 3, there is no clear FL algorithm that has universally superior performance in terms of convergence speed. For CIFAR100 and $K = 10$, FedGBO with SGDm converges fastest, whereas for the Shakespeare, $K = 10$, Mimelite with SGDm converges fastest. Furthermore, for FEMNIST none of the tested algorithms provided significant benefit compared to FedAvg in both the $K = 10$ and $K = 50$ scenarios. The reason for this result is likely because FEMNIST is a relatively simple greyscale task, so more sophisticated optimisation techniques do not much help convergence. 

Therefore, when planning a real-world FL deployment, the potential benefit of adaptive-FL should be taken into account. If the learning task is relatively simple, additional hyperparameters (which then need to be tuned) and communicated data added by adaptive-FL algorithms may outweigh the performance gain. Some non-adaptive FL algorithms can show enhanced performance compared to FedAvg without adding to the communication or computation cost (e.g., FedMAX for CIFAR100). However, on the more challenging CIFAR100 and Shakespeare tasks, adaptive FL algorithms provide significant speedup compared to FedAvg.

Comparing Figs. 2 and 3, some datasets and algorithms benefited from increased local computation, whereas others were hindered. For CIFAR100 with RMSProp, the best accuracy that all algorithms could achieve was lowered for $K = 50$. For Shakespeare, almost all algorithms and optimisers were able to converge faster for $K = 50$, and the ordering of which algorithm achieved the best accuracy changed. The relative performance of adaptive-FL algorithms on the different benchmark datasets warrants further investigation in future works, but task complexity and number of local samples appear to be important factors.

Alongside this, the communication costs of the adaptive-FL algorithms should be taken into account. We provide a more detailed analysis of the results considering final model accuracy, upload cost, and total computation in the Section 5.2. 
\\

\noindent \textbf{Choice of optimiser:} As shown in Figure 2, there was also no universally-best optimiser for each dataset and adaptive-FL algorithm, which matches the finding for centralised training \cite{DescendValley}. For example, the fastest convergence for Shakespeare, $K =10$ was using Mimelite with SGDm, whereas for Sent140 it was RMSProp and Adam in both $K=10$ and $K = 50$ scenarios. 

In both scenarios for Sent140, SGDm provided little improvement compared to FedAvg. The gradients computed in the Sent140 task are sparse (due to very sparse features). Adaptive learning-rate methods like RMSProp and Adam are known to provide good convergence rates for sparse-gradient tasks \cite{ConvergeAdaptGradNonconvex}, whereas the very noisy convergence of SGDm is due to the learning rate having to be set very large for this task (see hyperparameter table in code repository). For the $K = 50$ scenario, the learning rate could be set lower, resulting in slightly less erratic convergence and narrower confidence intervals. 

On the other hand, the rest of the tasks do not have sparse features, and RMSProp performed largely the worst. Therefore, these results suggest that when choosing an adaptive-FL optimiser, a good initial choice of optimiser is the one that works best on the task in the standard centralised setting. Also, as shown in Table 2, optimiser choice impacts the communication cost of FedGBO, MFL and Mimelite: for RMSProp and SGDm, the total download cost is doubled, and tripled for Adam. The relationship between convergence speed gain due to optimiser choice and increased communication cost represents another design consideration for FL engineers.
\\

\noindent \textbf{Combining strategies:} The adaptive-FL algorithms studied (FedGBO, Mimelite, MFL, AdaptiveFL) add improve the convergence rate of the global model through adaptive optimisation. However, in some cases simply modifying the local objective can result in faster convergence (e.g., FedMAX on CIFAR100). The adaptive-FL algorithms could readily be combined with objective-modifying algorithms, which may provide even greater performance without additional overheads. This combination presents an interesting avenue for future research.
\color{black}

\subsection{Considering Convergence, Communication and Computation}
There are many factors to consider when designing an FL algorithm for real-world use. Alongside convergence speed, the total computational cost of FL is very important. In cross-device FL, clients may be a highly diverse set of devices with a wide range of computational power (e.g., different generations of smartphones) \cite{AdvancesInFL}. This has two practical implications: the time taken to perform local training may be the most significant bottleneck \cite{TowardsEffSched}, and FL deployments that drop stragglers \cite{FLAtScale} may end up regularly dropping the same clients, skewing the global model in favour of faster clients. Furthermore, research indicates the total energy consumed by FL can exceed the energy consumed by centralised training, motivating FL algorithms  with lower computational costs \cite{FLSavePlanet}.

As well as model convergence and computation cost, other important factors include the total data communicated during FL (especially uploaded data considering the asymmetric bandwidth of the network edge), number of algorithm hyperparameters, and more. It is therefore difficult to compare FL algorithms when considering all these factors. In Table 4, we attempt to compare the tested algorithms using several factors by combining the per-round communication costs from Table 1, the convergence results from Fig. 2 and the computation costs of each algorithm for the $K = 10$ scenario using SGDm. Details about calculating the computation costs of the FL algorithms are given in Appendix B.

The left-hand section of Table 4 shows the maximum accuracy of each algorithm, and the total upload cost and FLOPs required to reach it (number of rounds multiplied by per-round upload/FLOPs). The right-hand section shows the total data and FLOPs required to match the maximum FedAvg accuracy for that scenario, giving an indication of convergence speed.

\begingroup
\setlength{\tabcolsep}{3pt}
\begin{table}[h]
\centering
\caption{Experimental results for different algorithms on 4 benchmark FL datasets. The left section displays the maximum accuracy achieved by each algorithm (95\% confidence intervals in brackets), and the total upload cost and FLOPs to achieve that accuracy. The right section shows the number of rounds taken to match FedAvg's accuracy for the given scenario, with the corresponding upload cost and FLOPs. Adaptive-FL algorithms use SGDm.}
\color{black}
\begin{tabular}{c | c c c | c c }
	\multirow{2}{*}{Algorithm} & Max Acc & Upload & FLOPs & Upload & FLOPs \\
						& (\%) & (GB) & ($\times 10^{12}$) & (GB) & ($\times 10^{12}$) \\
\hline
\multicolumn{6}{c}{CIFAR100} \\
\hline
	FedAvg   & 40.2 ($\pm 0.9$) & 112 & 120 & -    & -    \\
	FedProx  & 40.6 ($\pm 0.7$) & 115 & 124 & 115  & 124  \\
	FedMAX   & 45.6 ($\pm 1.0$) & $\bm{112}$ & $\bm{120}$ & 65.2 & 69.7 \\
	AFO      & 41.7 ($\pm 0.8$)& 112 & 120 & 59.4 & 63.5 \\
	MFL      & 47.7 ($\pm 0.3$) & 231 & 125 & 126  & 68.3 \\
	Mimelite & 46.8 ($\pm 0.4$) & 231 & 163 & 130  & 92.0 \\
	FedGBO   & $\bm{49.2}$ ($\pm 0.7$) & 115 & 124 & $\bm{22.1}$ & $\bm{23.8}$ \\
\hline
\multicolumn{6}{c}{Sent140} \\
\hline
	FedAvg   & 75.2 ($\pm 0.4$) & 3.9 & 0.27 & -   & -    \\
	FedProx  & 73.3 ($\pm 0.4$) & 4.0 & 0.31 & -   & -    \\
	AFO      & 75.0 ($\pm 1.0$) & $\bm{3.4}$ & $\bm{0.24}$ & -   & -    \\
	MFL      & 75.4 ($\pm 0.8$) & 6.1 & 0.26 & 6.1 & 0.26 \\
	Mimelite & $\bm{75.0}$ ($\pm 2.0$) & 6.9 & 0.32 & -   & -    \\
	FedGBO   & $\bm{76.5}$ ($\pm 0.2$) & $\bm{3.4}$ & 0.27 & $\bm{1.2}$ & $\bm{0.09}$ \\
\hline
\multicolumn{6}{c}{FEMNIST} \\
\hline
	FedAvg   & 86.6 ($\pm 0.2$) & $\bm{423}$ & $\bm{416}$ & - & - \\
	FedProx  & 86.6 ($\pm 0.2$) & 472 & 469 & 423 & 420 \\
	FedMAX   & 86.7 ($\pm 0.2$) & 472 & 464 & 423 & 416 \\
	AFO      & 86.4 ($\pm 0.6$) & 433 & 426 & 389 & 383 \\
	MFL      & $\bm{87.0}$ ($\pm 0.1$) & 846 & 423 & 418 & 209 \\
	Mimelite & $\bm{87.3}$ ($\pm 0.2$) & 720 & 544 & $\bm{282}$ & $\bm{213}$ \\
	FedGBO   & 86.4 ($\pm 0.3$) & 447 & 444 & 447 & 444 \\
\hline
\multicolumn{6}{c}{Shakespeare} \\
\hline
	FedAvg   & 53.9 ($\pm 0.1$) & 21.0 & 409  & - & -  \\
	FedProx  & 54.4 ($\pm 0.1$) & $\bm{20.4}$ & 396  & 13.0 & 396  \\
	FedMAX   & 54.6 ($\pm 0.1$) & 21.0 & 409  & 12.1 & 235  \\
	AFO      & 54.1 ($\pm 0.1$) & 21.0 & 409  & 16.1 & 314  \\
	MFL      & 53.9 ($\pm 0.2$) & 37.8 & $\bm{368}$  & -    & -    \\
	Mimelite & $\bm{56.6}$ ($\pm 0.3$) & 40.8 & 7325 & $\bm{7.6}$  & 1373 \\
	FedGBO   & 54.6 ($\pm 0.3$) & 20.6 & 401  & 8.9   & $\bm{173}$ \\
\end{tabular}
\end{table}
\endgroup

Table 4 shows that for CIFAR100, FedGBO achieved the highest accuracy whilst also having near-lowest uploaded data and total FLOPs. To match FedAvg's accuracy, FedGBO had by far the lowest upload cost and FLOPs, much lower than FedAvg. Similar performance is shown with Sent140. For FEMNIST, FedGBO had competitive performance in terms of model accuracy, upload cost and FLOPs, and for Shakespeare, had competitive accuracy whilst having among the lowest upload cost and computation. We therefore believe that FedGBO shows a good trade-off between convergence rate, final accuracy, uploaded data, and computational cost. 

Mimelite and FedGBO both use fixed global optimisers. However, the computation performed by Mimelite is much  higher than FedGBO due to computing full-batch gradients (see Appendix B). This motivates the question of whether computing these full-batch gradients is a good use of local computation, especially considering the results from the Shakespeare scenario (where Mimelite achieved the highest model accuracy but with more than $10\times$ the total computation). To test this, we modified Mimelite to use minibatch unbiased gradients instead (which we call MimeXlite: Mime-`extra'-light). MimeXlite has the same computational cost as FedGBO, but retains the $2\times$ upload cost.
\color{black}

\begingroup
\setlength{\tabcolsep}{3pt}
\begin{table}[h]
\centering
\caption{Maximum accuracy achieved by FedGBO, Mimelite and MimeXlite on the Shakespeare dataset (95\% confidence intervals given in brackets).}
\color{black}
\begin{tabular}{c c | c c c}
	& Dataset & FedGBO & Mimelite & MimeXlite \\
\hline
	\multirow{3}{*}{K = 10} & RMSProp & 55.4 ($\pm 0.1$) & 54.8 ($\pm 0.2$) & 55.9 ($\pm 0.3$) \\
							& SGDm    & 54.6 ($\pm 0.3$) & 56.6 ($\pm 0.3$) & 51.9 ($\pm 0.3$) \\
							& Adam    & 55.6 ($\pm 0.1$) & 56.6 ($\pm 0.2$) & 52.8 ($\pm 0.2$) \\
\hline
	\multirow{3}{*}{K = 50} & RMSProp & 56.8 ($\pm 0.1$) & 56.7 ($\pm 0.2$) & 55.9 ($\pm 0.1$) \\
							& SGDm    & 57.1 ($\pm 0.1$) & 57.2 ($\pm 0.2$) & 50.4 ($\pm 0.3$) \\
							& Adam    & 56.4 ($\pm 0.2$) & 56.2 ($\pm 0.1$) & 52.5 ($\pm 0.2$) \\
\end{tabular}
\end{table}
\endgroup

Table 5 shows that the performance of MimeXlite is significantly worse than FedGBO or Mimelite in almost all Shakespeare scenarios. The performance drop is worst for SGDm and Adam. This is likely due to the high variance of minibatch gradients harming the momentum parameters (as used in SGDm and Adam), but having a smaller impact on RMSProp. We believe FedGBO `gets away' with computing only minibatch gradients because the optimiser gradients are averaged over the local steps, helping to lower the variance of the global optimiser gradients (despite the gradients being biased).
\color{black}

\section{Conclusion}
\noindent In this work, we proposed the Federated Global Biased Optimiser (FedGBO) algorithm that incorporates a set of global statistics for a generic machine learning optimiser within the Federated Learning (FL) process. FedGBO demonstrates fast convergence rates, whilst having lower communication and computation costs compared to other state-of-the-art FL algorithms that use adaptive optimisation. We showed that FedGBO with a generic optimiser can be formulated as centralised training using a perturbed gradient and optimiser update, allowing analysis of generic centralised optimisers to be extended to FedGBO. We achieve the same convergence rate for FedGBO with SGDm as a recent analysis on centralised SGDm does, plus an extra FL-related error term that decays with $\eta^2$. We then performed an extensive set of comparison experiments using 6 competing FL algorithms (3 of which also use adaptive optimisation) and 3 different optimisers on 4 benchmark FL datasets. These experiments highlighted FedGBO's highly competitive performance, especially considering FedGBO's low computation and communication costs, and provided practical insights into the choice of adaptive-FL algorithms and optimisers, and communication and computation trade-offs within FL.

\bibliographystyle{ieee}
\bibliography{references}

\begin{thebibliography}{10}
\providecommand{\url}[1]{#1}
\csname url@samestyle\endcsname
\providecommand{\newblock}{\relax}
\providecommand{\bibinfo}[2]{#2}
\providecommand{\BIBentrySTDinterwordspacing}{\spaceskip=0pt\relax}
\providecommand{\BIBentryALTinterwordstretchfactor}{4}
\providecommand{\BIBentryALTinterwordspacing}{\spaceskip=\fontdimen2\font plus
\BIBentryALTinterwordstretchfactor\fontdimen3\font minus
  \fontdimen4\font\relax}
\providecommand{\BIBforeignlanguage}[2]{{%
\expandafter\ifx\csname l@#1\endcsname\relax
\typeout{** WARNING: IEEEtran.bst: No hyphenation pattern has been}%
\typeout{** loaded for the language `#1'. Using the pattern for}%
\typeout{** the default language instead.}%
\else
\language=\csname l@#1\endcsname
\fi
#2}}
\providecommand{\BIBdecl}{\relax}
\BIBdecl

\bibitem{AdvancesInFL}
P.~Kairouz, H.~B. McMahan, B.~Avent, A.~Bellet, M.~Bennis, A.~N. Bhagoji, K.~A.
  Bonawitz, Z.~Charles \emph{et~al.}, ``Advances and open problems in federated
  learning,'' \emph{Foundations and Trends in Machine Learning}, vol.~14, no.
  1–2, pp. 1--210, 2021.

\bibitem{FedAvgPaper}
B.~McMahan, E.~Moore, D.~Ramage, and B.~A. y~Arcas, ``Communication-efficient
  learning of deep networks from decentralized data,'' in \emph{Proc.
  International Conference on Artifical Intelligence and Statistics (AISTATS)},
  vol.~54, 2017, pp. 1273--1282.

\bibitem{AdaptiveFedOpt}
S.~J. Reddi, Z.~Charles, M.~Zaheer, Z.~Garrett, K.~Rush, J.~Kone{\v{c}}n{\'y},
  S.~Kumar, and H.~B. McMahan, ``Adaptive federated optimization,'' in
  \emph{Proc. International Conference on Learning Representations (ICLR)},
  2021.

\bibitem{SCAFFOLD}
S.~P. Karimireddy, S.~Kale, M.~Mohri, S.~Reddi, S.~Stich, and A.~T. Suresh,
  ``{SCAFFOLD}: Stochastic controlled averaging for federated learning,'' in
  \emph{Proc. International Conference on Machine Learning (ICML)}, vol. 119,
  2020, pp. 5132--5143.

\bibitem{FLAtScale}
K.~Bonawitz, H.~Eichner, W.~Grieskamp, D.~Huba, A.~Ingerman, V.~Ivanov,
  C.~Kiddon, J.~Kone{\v{c}}n{\'y} \emph{et~al.}, ``Towards federated learning
  at scale: System design,'' in \emph{Proc. Conference on Machine Learning and
  Systems (SysML)}, 2019.

\bibitem{FedProx}
T.~Li, A.~K. Sahu, M.~Zaheer, M.~Sanjabi, A.~Talwalkar, and V.~Smith,
  ``Federated optimization in heterogeneous networks,'' in \emph{Proc. Machine
  Learning and Systems (MLSys)}, vol.~2, pp. 429--450, 2020.

\bibitem{FedMAX}
W.~Chen, K.~Bhardwaj, and R.~Marculescu, ``Fedmax: mitigating activation
  divergence for accurate and communication-efficient federated learning,'' in
  \emph{Joint European Conference on Machine Learning and Knowledge Discovery
  in Databases}, 2020, pp. 348--363.

\bibitem{MFL}
W.~Liu, L.~Chen, Y.~Chen, and W.~Zhang, ``Accelerating {Federated} {Learning}
  via {Momentum} {Gradient} {Descent},'' \emph{IEEE Transactions on Parallel
  and Distributed Systems}, vol.~31, no.~8, pp. 1754--1766, aug 2020.

\bibitem{Mime}
S.~P. Karimireddy, M.~Jaggi, S.~Kale, M.~Mohri, S.~J. Reddi, S.~U. Stich, and
  A.~T. Suresh, ``Mime: Mimicking centralized stochastic algorithms in
  federated learning,'' \emph{arXiv e-prints arXiv:2008.03606}, 2020.

\bibitem{SAFA}
W.~Wu, L.~He, W.~Lin, R.~Mao, C.~Maple, and S.~Jarvis, ``Safa: A
  semi-asynchronous protocol for fast federated learning with low overhead,''
  \emph{IEEE Transactions on Computers}, vol.~70, no.~5, pp. 655--668, 2021.

\bibitem{BAFL}
L.~Feng, Y.~Zhao, S.~Guo, X.~Qiu, W.~Li, and P.~Yu, ``Blockchain-based
  asynchronous federated learning for internet of things,'' \emph{IEEE
  Transactions on Computers}, 2021.

\bibitem{FedForest}
Y.~Liu, Y.~Liu, Z.~Liu, Y.~Liang, C.~Meng, J.~Zhang, and Y.~Zheng, ``Federated
  forest,'' \emph{IEEE Transactions on Big Data}, pp. 1--1, 2020.

\bibitem{FedMAB}
C.~Shi, C.~Shen, and J.~Yang, ``Federated multi-armed bandits with
  personalization,'' in \emph{Proc. International Conference on Artificial
  Intelligence and Statistics (AISTATS)}, vol. 130, 2021, pp. 2917--2925.

\bibitem{FedDyn}
D.~A.~E. Acar, Y.~Zhao, R.~Matas, M.~Mattina, P.~Whatmough, and V.~Saligrama,
  ``Federated learning based on dynamic regularization,'' in \emph{Proc.
  International Conference on Learning Representations (ICLR)}, 2021.

\bibitem{ConvAccTradeOffs}
Z.~Charles and J.~Kone\v{c}n\'y, ``Convergence and accuracy trade-offs in
  federated learning and meta-learning,'' in \emph{Proc. International
  Conference on Artificial Intelligence and Statistics (AISTATS)}, vol. 130,
  2021, pp. 2575--2583.

\bibitem{LocalFixedPoint}
G.~Malinovskiy, D.~Kovalev, E.~Gasanov, L.~Condat, and P.~Richtarik, ``From
  local {SGD} to local fixed-point methods for federated learning,'' in
  \emph{Proc. International Conference on Machine Learning (ICML)}, 2020, pp.
  6692--6701.

\bibitem{IsLocalSGDBetter}
B.~Woodworth, K.~K. Patel, S.~Stich, Z.~Dai, B.~Bullins, B.~Mcmahan, O.~Shamir,
  and N.~Srebro, ``Is local {SGD} better than minibatch {SGD}?'' in \emph{Proc.
  International Conference on Machine Learning (ICML)}, vol. 119, 2020, pp.
  10\,334--10\,343.

\bibitem{AdaptFLResourceConstraint}
J.~Zhang, S.~Guo, Z.~Qu, D.~Zeng, Y.~Zhan, Q.~Liu, and R.~A. Akerkar,
  ``Adaptive federated learning on non-iid data with resource constraint,''
  \emph{IEEE Transactions on Computers}, 2021.

\bibitem{SGVR}
R.~Johnson and T.~Zhang, ``Accelerating stochastic gradient descent using
  predictive variance reduction,'' \emph{Advances in neural information
  processing systems}, vol.~26, pp. 315--323, 2013.

\bibitem{FLKeywordSpotting}
D.~Leroy, A.~Coucke, T.~Lavril, T.~Gisselbrecht, and J.~Dureau, ``Federated
  learning for keyword spotting,'' in \emph{Proc. International Conference on
  Acoustics, Speech and Signal Processing (ICASSP)}, 2019, pp. 6341--6345.

\bibitem{CommEffFLIoT}
J.~Mills, J.~Hu, and G.~Min, ``Communication-efficient federated learning for
  wireless edge intelligence in {IoT},'' \emph{IEEE Internet of Things
  Journal}, vol.~7, no.~7, pp. 5986--5994, 2020.

\bibitem{ClientSideOptStrats}
------, ``Client-side optimisation strategies for communication-efficient
  federated learning,'' \emph{IEEE Communications Magazine}, pp. 1--11, 2022.

\bibitem{RMSProp}
T.~Tieleman and G.~Hinton, ``Rmsprop, coursera: Neural netwroks for machine
  learning,'' \emph{Technical Report}, 2012.

\bibitem{Adam}
D.~P. Kingma and J.~Ba, ``Adam: {A} method for stochastic optimization,'' in
  \emph{Proc. International Conference on Learning Representations (ICLR)},
  2015.

\bibitem{SimpleConvergeAdam}
A.~D\'{e}fossez, L.~Bottou, F.~Bach, and N.~Usunier, ``A simple convergence
  proof of adam and adagrad,'' \emph{arXiv e-prints arXiv:2003.02395}, 2020.

\bibitem{LEAF}
S.~Caldas, P.~Wu, T.~Li, J.~Konecn{\'{y}}, H.~B. McMahan, V.~Smith, and
  A.~Talwalkar, ``Leaf: A benchmark for federated settings,'' in \emph{NeurIPS
  Workshop on Federated Learnign for Data Privacy and Confidentiality}, 2019.

\bibitem{DescendValley}
R.~Schmidt, F.~Schneider, and P.~Hennig, ``Descending through a crowded valley
  - benchmarking deep learning optimizers,'' in \emph{Proc. International
  Conference on Machine Learning (ICML)}, vol. 139, 2021, pp. 9367--9376.

\bibitem{ConvergeAdaptGradNonconvex}
D.~Zhou, J.~Chen, Y.~Cao, Y.~Tang, Z.~Yang, and Q.~Gu, ``On the convergence of
  adaptive gradient methods for nonconvex optimization,'' in \emph{NeurIPS
  Workshop on Optimization for Machine Learning (OPT)}, 2020.

\bibitem{TowardsEffSched}
C.~Wang, Y.~Yang, and P.~Zhou, ``Towards efficient scheduling of federated
  mobile devices under computational and statistical heterogeneity,''
  \emph{IEEE Transactions on Parallel and Distributed Systems}, vol.~32, no.~2,
  pp. 394--410, 2021.

\bibitem{FLSavePlanet}
X.~Qiu, T.~Parcollet, D.~J. Beutel, T.~Topal, A.~Mathur, and N.~D. Lane, ``Can
  federated learning save the planet?'' in \emph{NeurIPS Workshop on Tackling
  Climate Change with Machine Learning}, 2020.

\bibitem{MeasureAlgEff}
D.~Hernandez and T.~B. Brown, ``Measuring the algorithmic efficiency of neural
  networks,'' \emph{arXiv e-prints arXiv:2005.04305}, 2020.

\end{thebibliography}

\clearpage

\appendices
\section{Proof of Theorems}
In the proofs below, unless otherwise specified, the expectation $\E[\cdot]$ is over both the clients selected randomly in each round of FedGBO, and the random sampling of minibatches in local updates. 
\\

\noindent \textbf{Lemma 1} (Bounding the momentum term). \textit{Given $\tilde{\eta} = \eta K > 0$, $0 \leq \beta < 1$, $\bm{m}_t$ defined in (4) and (5), and Assumptions A1-A5, then}
\begin{equation*}
\E \left[ \| \bm{m}_t \|^2 \right] \leq 2 \frac{R^2 + \tilde{\eta}^2 Y}{(1 - \beta)^2},
\end{equation*}
where $Y = 18 B^2 L^2 \left( R^2 + G^2 + \frac{\sigma^2}{K} \right)$.

\begin{proof} Taking any iteration $t$:
\begin{align*}
\E \left[ \| \bm{m}_t \|^2 \right] 	&= \E \left[ \left\| \sum_{i = 0}^{t - 1} \beta^i (\bm{g}_{t-i} + \bm{e}_{t - i}) \right\|^2 \right] \\
								&\leq \left( \sum_{i = 0}^{t - 1} \beta^i \right) \sum_{i = 0}^{t - 1} \beta^i \E \left[ \| \bm{g}_{t - i} + \bm{e}_{t - i} \|^2 \right] \\
								&\leq \frac{1}{1 - \beta} \sum_{i = 0}^{t - 1} \beta^i \E \left[ \| \bm{g}_{t - i} + \bm{e}_{t - i} \|^2 \right] \\
								&\leq \frac{2}{1 - \beta} \sum_{i = 0}^{t - 1} \beta^i \left( \E \left[ \| \bm{g}_{t - i} \|^2 \right] + \E \left[ \| \bm{e}_{t - i} \|^2 \right] \right) \\
								&\leq 2 \frac{\E \left[ \| \bm{g}_{t - i} \|^2 \right] + \E \left[ \| \bm{e}_{t - i} \|^2 \right]}{(1 - \beta)^2} \\
								&\leq 2 \frac{R^2 + \tilde{\eta}^2 Y}{(1 - \beta)^2}. \qedhere \\
\end{align*}
\end{proof}
\noindent Here, the first inequality comes from Jensen, the third inequality is from Cauchy-Schwarz and the linearity of expectation, and the last inequality from Assumption 3 and our definition of $Y$.
\\

\noindent \textbf{Lemma 2} (bounding the negative term of the Descent Lemma (16)). \textit{Given $\tilde{\eta} = \eta K > 0$, $0 \leq \beta < 1$, $\bm{m}_t$ defined in (4) and (5), and Assumptions A1-A5, then}
\begin{multline*}
\E [\nabla F(\bm{x}_{t - 1})^\top \bm{m}_{t}] \geq \frac{1}{2} \sum_{k = 0}^{t - 1} \beta^k \E \left[ \|\nabla F(\bm{x}_{t-k-1})\|^2 \right] \\ 
- \frac{\tilde{\eta}^2 Y}{2(1 - \beta)} - \frac{2 \tilde{\eta} L \beta (R^2 + \tilde{\eta}^2 Y)}{(1 - \beta)^3} .
\end{multline*}
where $Y = 18 B^2 L^2 \left( R^2 + G^2 + \frac{\sigma^2}{K} \right)$.

\begin{proof} To ease notation, we denote $\bm{G}_t = \nabla F(\bm{x}_{t-1})$ and $\bm{c}_t = \bm{g}_t + \bm{e}_t$:
\begin{align}
\bm{G}_t^\top \bm{m}_t &= \sum_{k=0}^{t-1} \beta^k \bm{G}_t^\top \bm{c}_t \nonumber \\
					&= \sum_{k=0}^{t-1} \beta^k \bm{G}_{t-k}^\top \bm{c}_t + \sum_{k=0}^{t-1} \beta^k (\bm{G}_t - \bm{G}_{t-k})^\top \bm{c}_t .
\end{align}
Due to $F$ being $L$-smooth (Assumption 4), and using the relaxed triangle inequality:
\begin{align}
\| \bm{G}_t - \bm{G}_{t - k} \|^2 	&\leq L^2 \left\| \sum_{l = 1}^{k} \tilde{\eta} \bm{m}_{n-l} \right\|^2 \nonumber \\ 
									&\leq \tilde{\eta}^2 L^2 k \sum_{l = 1}^{k} \| \bm{m}_{n - l} \|^2 .
\end{align}
Using $(\lambda \bm{x} - \bm{y})^2 \geq 0, \forall \bm{x}, \bm{y} \in \mathbb{R}^d, \lambda > 0$, and hence $\| \bm{x}\bm{y} \| \leq \frac{\lambda}{2}\|\bm{x}\|^2 + \frac{1}{2 \lambda}\| \bm{y} \|^2$, we can use $\bm{x} = G_t - G_{t - k}$, $\bm{y} = \bm{c}_t$, $\lambda = \frac{1 - \beta}{k \tilde{\eta}L}$, and substitute this inequality into (8):
\begin{multline*}
\bm{G}_t^\top \bm{m}_t \geq \sum_{k = 0}^{t - 1} \beta^k \bm{G}_{t - k}^\top \bm{c}_{t - k} \\ 
- \sum_{k = 1}^{t - 1} \frac{\beta^k}{2} \Bigg( \bigg( (1 - \beta) \tilde{\eta} L \sum_{l = 1}^k \| \bm{m}_{t - l} \|^2 \bigg) \\
 + \frac{\tilde{\eta} L k}{1 - \beta}\| \bm{c}_{t - k} \|^2 \Bigg).
\end{multline*}
Taking the expectation of both sides gives:
\begin{multline}
\E[ \bm{G}_t^\top \bm{m}_t] \geq \sum_{k = 0}^{t - 1} \beta^k \E[ \bm{G}_{t - k}^\top \bm{c}_{t - k}] \\ 
- \tilde{\eta} L \sum_{k = 1}^{t - 1} \frac{\beta^k}{2} \Bigg( \bigg( (1 - \beta) \sum_{l = 1}^k  \E [ \| \bm{m}_{t - l} \|^2 ] \bigg) \\
+ \frac{k}{1 - \beta} \E [ \| \bm{c}_{t - k} \|^2 ] \Bigg).
\end{multline}
Bounding the $\E[ \bm{G}_{t - k}^\top \bm{c}_{t - k}]$ term, this time using $(\bm{G}_{t - k}^\top + \bm{e}_{t - k})^2 \geq 0$, and hence $\bm{G}_{t - k}^\top\bm{e}_{t - k} \geq -\frac{1}{2}( \| \bm{G}_{t - k}\|^2 + \| \bm{e}_{t - k} \|^2)$, and linearity of expectation:
\begin{align}
\E[\bm{G}_{t - k}^\top \bm{c}_{t - k}] 	&= \E \left[ \bm{G}_{t - k}^\top (\bm{g}_{t - k} + \bm{e}_{t - k}) \right] \nonumber \\
								&= \E \left[ \bm{G}_{t - k}^\top \bm{g}_{t - k} \right] + \E \left[ \bm{G}_{t - k}^\top \bm{e}_{t - k} \right] \nonumber \\
								&\geq \E \left[ \| \bm{G}_{t - k}\|^2 \right] - \frac{1}{2} \E \left[ \| \bm{G}_{t - k} \|^2 + \| \bm{e}_{t - k} \| ^2 \right] \nonumber \\
								&= \frac{1}{2} \left( \E \left[ \| \bm{G}_{t - k}\|^2 \right] - \E \left[ \|\bm{e}_{t-k}\|^2 \right] \right).
\end{align}
Using Cauchy-Schwarz, linearity of expectation, (3), Assumption 3, and the definition of $Y$:
\begin{align}
\E[\|\bm{c}_{t - k}\|^2] 	&= \E[\|\bm{g}_{t-k} + \bm{e}_{t-k} \|^2] \nonumber \\
							&\leq 2( \E[\|\bm{g}_{t-k}\|^2] + \E[\|\bm{e}_{t-k}\|^2] ) \nonumber\\
							&\leq 2(R^2 + \tilde{\eta}^2 Y).
\end{align}
Then inserting Lemma 1, (11) and (12) into (10) gives:
\begin{align}
\E[\bm{G}_{t}^\top \bm{m}_t] &\geq \sum_{k=0}^{t} \frac{\beta^k}{2} \left( \E[\| \bm{G}_{t-k}\|^2] - \E[\|\bm{e}_{t-k}\|^2] \right)  \nonumber \\
& \qquad - \tilde{\eta} L \sum_{k = 1}^{t - 1} \frac{\beta^k}{2} \Bigg( \bigg( (1 - \beta) \sum_{l = 1}^k  2 \frac{R^2 + \tilde{\eta}^2 Y}{(1 - \beta)^2} \bigg) \nonumber \\
& \qquad \qquad \qquad \qquad \qquad + \frac{k}{1 - \beta} 2(R^2 + \tilde{\eta}^2 Y) \Bigg) \nonumber \\
					&= \sum_{k=0}^{t} \frac{\beta^k}{2} \E[\| \bm{G}_{t-k}\|^2] - \sum_{k=0}^{t} \frac{\beta^k}{2} \E[\|\bm{e}_{t-k}\|^2] \nonumber \\ 
					& \qquad - \tilde{\eta} L (R^2 + \tilde{\eta}^2 Y) \sum_{k = 1}^{t - 1} \beta^k \Bigg( \bigg( \sum_{l = 1}^k  \frac{1}{(1 - \beta)} \bigg) \nonumber \\
					& \hspace*{17em} + \frac{k}{1 - \beta} \Bigg) \nonumber \\
					&= \sum_{k=0}^{t} \frac{\beta^k}{2} \E[\| \bm{G}_{t-k}\|^2] - \frac{\tilde{\eta}^2 Y}{2(1 - \beta)} \nonumber \\
					& \qquad - \frac{2\tilde{\eta} L}{1 - \beta} (R^2 + \tilde{\eta}^2 Y) \sum_{k = 1}^{t - 1} \beta^k k.
\end{align}
Using Lemma B.2 from [2], which states that, for $0 < a < 1, i \in \mathbb{N}, Q \geq i$:
\begin{equation}
\sum_{q=i}^Q a^q q \leq \frac{a}{(1 - a)^2},
\end{equation}
then (13) becomes:
\begin{multline}
\E[ \bm{G}_{t}^\top \bm{m}_t] \geq \frac{1}{2} \sum_{k = 0}^{t - 1} \beta^k \E[\| \bm{G}_{t-k}\|^2] - \frac{\tilde{\eta}^2 Y}{2(1 - \beta)} \\
- \frac{2 \tilde{\eta} L \beta (R^2 + \tilde{\eta}^2 Y)}{(1 - \beta)^3} .
\end{multline}
Substituting the definition of $\bm{G}_t$ completes the Lemma.
\end{proof}

\noindent \textbf{Theorem 1} (Convergence of FedGBO using SGDm) \textit{Let the assumptions A1-A5 hold, the total number of communication rounds $T > (1 - \beta)^{-1}$, $\tilde{\eta} = K\eta > 0, 0 \leq \beta < 1$, and using the update steps given by (4) and (5), then:}
\begin{multline*}
\E[\|\nabla F(\bm{x}_t)\|^2] \leq \frac{2(1 - \beta)}{\tilde{\eta} \tilde{T}}(F(\bm{x}_0) - F^*) + \frac{T \tilde{\eta}^2 Y}{\tilde{T}} \\
+ \frac{2T\tilde{\eta}L(1+\beta)(R^2 + \tilde{\eta}^2 Y)}{\tilde{T}(1 - \beta)},
\end{multline*}
where $\tilde{T} = T - \frac{\beta}{1 - \beta}$, and $Y = 18 B^2 L^2 \left( R^2 + G^2 + \frac{\sigma^2}{K} \right)$. 

\begin{proof} Using the $L$-smoothness of $F$, and the update rules from (4) and (5), the Descent Lemma of FedGBO with SGDm is given by:
\begin{equation}
F(\bm{x}_t) \leq F(\bm{x}_{t-1}) - \tilde{\eta} \bm{G}_t^\top \bm{m}_t + \frac{\tilde{\eta}^2 L}{2} \| \bm{m}_t \|^2 .
\end{equation}
Taking expectations of both sides, and inserting Lemma 1 and Lemma 2 into (16):
\begin{align}
\E[F(\bm{x}_t)] &\leq \E[F(\bm{x}_{t - 1})] - \frac{\tilde{\eta}}{2} \sum_{k=0}^{t-1} \beta^k \E[\|\bm{G}_{t-k}\|^2] \nonumber \\
& \qquad + \frac{\tilde{\eta}^3 Y}{2(1 - \beta)} + \frac{2 \tilde{\eta}^2 L \beta (R^2 + \tilde{\eta}^2 Y)}{(1 - \beta)^3} \nonumber \\
& \qquad + \frac{\tilde{\eta}^2 L(R^2 + \tilde{\eta}^2 Y)}{(1 - \beta)^2} \nonumber \\
				&= \E[F(\bm{x}_{t - 1})] - \frac{\tilde{\eta}}{2} \sum_{k=0}^{t-1} \beta^k \E[\|\bm{G}_{t-k}\|^2] \nonumber \\
				& \qquad + \frac{\tilde{\eta}^3 Y}{2(1 - \beta)} \nonumber \\
				& \qquad + \frac{\tilde{\eta}^2 L (1 + \beta)(R^2 + \tilde{\eta}^2 Y)}{(1 - \beta)^3}.
\end{align}
Rearranging (17), summing over the iterations $t = 1 \cdots T$ and telescoping:
\begin{multline}
\frac{\tilde{\eta}}{2} \sum_{t=1}^{T} \sum_{k = 0}^{t - 1} \beta^k \E \left[ \|\bm{G}_{t-k}\|^2 \right] \leq F(\bm{x}_0) - \E \left[ F(\bm{x}_T) \right] + \frac{\tilde{\eta}^3 T Y}{2(1 - \beta)} \\
+ \frac{\tilde{\eta}^2 L T (1 + \beta) (R^2 + \tilde{\eta}^2 Y)}{(1 - \beta)^3}.
\end{multline}
We proceed as [2] do to bound the left hand side of (18). We introduce the change of index $i = t - k$:
\begin{align}
\frac{\tilde{\eta}}{2} \sum_{t=1}^{T} \sum_{k = 0}^{t - 1} \beta^k \E \left[ \|\bm{G}_{t-k}\|^2 \right] &= \frac{\tilde{\eta}}{2} \sum_{t=1}^{T} \sum_{i = 0}^{t} \beta^{t-i} \E \left[ \|\bm{G}_{i}\|^2 \right] \nonumber \\
			&= \frac{\tilde{\eta}}{2} \sum_{i = 0}^{T} \E \left[ \|\bm{G}_{i}\|^2 \right] \sum_{t=1}^{T} \beta^{t-i} \nonumber \\
			&= \frac{\tilde{\eta}}{2 (1 - \beta)} \sum_{i=1}^{T} \bigg( \E\left[\| \nabla F(\bm{x}_{i - 1}) \|^2 \right] \nonumber \\
			& \hspace*{9em} (1 - \beta^{T-i+1}) \bigg) \nonumber \\
			&= \frac{\tilde{\eta}}{2 (1 - \beta)} \sum_{i=0}^{T-1} \bigg( \E\left[\| \nabla F(\bm{x}_{i}) \|^2 \right] \nonumber \\
			& \hspace*{9em} (1 - \beta^{T-i}) \bigg) .
\end{align}
(19) shows a non-normalised iteration probability as defined in Section 4.2 ($\mathbb{P}[t = j] \propto 1 - \beta^{T - j}$). The normalisation constant for summing to 1 is:
\begin{equation}
\sum_{i = 0}^{T - 1} 1 - \beta^{T - i} = T - \beta \frac{1 - \beta^T}{1 - \beta} \geq T - \frac{\beta}{1 - \beta} = \tilde{T}. 
\end{equation}
(19) can then be used in (20) to obtain:
\begin{equation}
\frac{\tilde{\eta}}{2} \sum_{t=1}^{T} \sum_{k = 0}^{t - 1} \beta^k \E \left[ \|\bm{G}_{t-k}\|^2 \right] \geq \frac{\tilde{\eta}\tilde{T}}{2(1 - \beta)} \E \left[\| \nabla F(\bm{x}_t) \|^2 \right].
\end{equation}

Inserting (21) into (18), and using the lower bound $F(\bm{x}) \geq F^*$ completes the proof:
\begin{multline*}
\E[\|\nabla F(\bm{x}_t)\|^2] \leq \frac{2(1 - \beta)}{\tilde{\eta} \tilde{T}}(F(\bm{x}_0) - F^*) + \frac{T \tilde{\eta}^2 Y}{\tilde{T}} \\
+ \frac{2T\tilde{\eta}L(1+\beta)(R^2 + \tilde{\eta}^2 Y)}{\tilde{T}(1 - \beta)^2}. \qedhere
\end{multline*}
\end{proof}

\section{Computation of FL Algorithms}
We calculated the total number of Floating Point Operations (FLOPs) required on average to compute the local update of the FL algorithms studied in this paper, for $K = 10$ and $K = 50$ local steps, shown in Tables 7 and 8 respectively. The total FLOPs were calculated as the sum of FLOPs required to compute $K$ minibatch gradients \cite{MeasureAlgEff} and update the local model using the algorithm's optimisation strategy. The Mimelite algorithm also requires computing full-batch gradients. We do not consider the cost of work on the server as it is represents a small fraction of the total. We only calculate FLOPs from an algorithmic standpoint, and do not consider optimisations such as vectorised operations.

The total FLOPs for a single forward-backward pass were computed using the Thop library (\url{https://github.com/Lyken17/pytorch-OpCounter/}), and the formula from the following OpenAI page: \url{https://openai.com/blog/ai-and-compute/}.

Local update steps involve computing gradients and applying them to the local model, plus any updates to a local optimiser if used. For example, for FedGBO using RMSProp, Table 1 shows that there are 5 operations per step that act on either $\bm{g}$ or $\bm{v}$: ($+$), ($\times -\eta$), ($\div$), ($\sqrt{\bm{v}}$), ($+ \epsilon$). MFL with RMSProp has 8 total operations: those 5 plus 3 to track the local $\bm{v}$ values. FedGBO, Mime and MFL all have 4 operations per SGDm update, as MFL's local tracking can be incorporated into the model update. FedAvg, AdaptiveFedOpt and FedMAX use vanilla SGD with 2 operations: ($+$), ($\times -\eta$). FedProx uses 5 operates due to the proximal update. Finally, Mimelite adds the cost of computing full-batch gradients. 

The formulas for computing the FLOPs for the local update of each algorithm, $N$, are therefore as follows:
\begin{align}
\mathrm{N}_{\mathrm{FedGBO}} 			&= K (B(\mathrm{fwd + bwd}) + C_{\mathrm{Fixed}}|\bm{x}|), \nonumber \\
\mathrm{N}_{\mathrm{Mimelite}} 			&= K (B(\mathrm{fwd + bwd}) + C_{\mathrm{Fixed}}|\bm{x}|) \nonumber \\
										& \qquad + n(\mathrm{fwd} + \mathrm{bwd}), \nonumber \\
\mathrm{N}_{\mathrm{MFL}} 				&= K (B(\mathrm{fwd + bwd}) + C_{\mathrm{Moving}}|\bm{x}|), \nonumber \\
\mathrm{N}_{\mathrm{AdaptiveFedOpt}} 	&= K (B(\mathrm{fwd + bwd}) + 2|\bm{x}|), \nonumber \\
\mathrm{N}_{\mathrm{SGD}} 				&= K (B(\mathrm{fwd + bwd}) + 2|\bm{x}|), \nonumber \\
\mathrm{N}_{\mathrm{FedProx}}			&= K (B(\mathrm{fwd + bwd}) + 5|\bm{x}|),
\end{align}
\noindent where $K$ is the number of local updates, $B$ is the minibatch size, $(\mathrm{fwd}+\mathrm{bwd})$ are the FLOPs required to compute the forward plus backward passes on one sample, $C_*$ is the number of operations for the type of local adaptive optimisation (see Table 6), $|\bm{x}|$ is the number of model parameters, and $n$ is the average number of client samples, as shown in Table 3.

\begin{table}[h]
\centering
\caption{Number of operations per local step of fixed or moving local optimisers.}
\begin{tabular}{ c | c c }
	Algorithm & $C_{\mathrm{Fixed}}$ & $C_{\mathrm{Moving}}$ \\ 
	\hline
	RMSProp	& 5 & 5 \\
	SGDm 	& 5 & 8 \\
	Adam 	& 8 & 11 \\
\end{tabular}
\end{table}

\begingroup
\begin{table}[h]
\setlength{\tabcolsep}{4pt} 
\centering
\caption{Total average FLOPs ($\times 10^8$) performed per client per round for different FL tasks, using different optimisers and FL algorithms: \colorbox{red}{FedGBO}, \colorbox{blue}{Mimelite}, \colorbox{green}{MFL} using RMSProp, SGDm and Adam. AdaptiveFedOpt, FedAvg and FedMAX use \colorbox{gray}{SGD} and \colorbox{purple}{FedProx}. Values shown for $K = 10$ local update steps, to 3 significant figures.}
\vspace{1mm}
\begin{tabular}{ c | c c c | c | c}

	Task & RMSProp & SGDm & Adam & SGD & FedProx \\ 
	\hline
	\multirow{3}{*}{CIFAR100} 		& \CR 50.2 & \CR 50.2 & \CR 50.6 & \CK  & \CPRPL\\
						      		& \CB 65.7 & \CB 65.7 & \CB 66.1 & \CK 49.8 & \CPRPL 50.2\\
   						      		& \CG 50.2 & \CG 50.6 & \CG 50.9 & \CK & \CPRPL \\
  	\hline
   						      		
	\multirow{3}{*}{Sent140} 		& \CR 0.0290 & \CR 0.0290 & \CR 0.0320 & \CK & \CPRPL \\
						      		& \CB 0.0350 & \CB 0.0350 & \CB 0.0365 & \CK 0.0260 & \CPRPL 0.0290 \\
   						      		& \CG 0.0290 & \CG 0.0320 & \CG 0.0350 & \CK & \CPRPL \\
  	\hline

   	\multirow{3}{*}{FEMNIST} 		& \CR 32.2 & \CR 32.2 & \CR 32.4 & \CK & \CPRPL \\
						      		& \CB 49.0 & \CB 49.0 & \CB 49.3 & \CK 31.9 & \CPRPL 32.2 \\
   						      		& \CG 32.2 & \CG 32.4 & \CG 32.7 & \CK & \CPRPL \\
  	\hline
   				      		
	\multirow{3}{*}{Shakespeare}	& \CR 118  & \CR 118  & \CR 118  & \CK & \CPRPL \\
						      		& \CB 2180 & \CB 2180 & \CB 2180 & \CK 118 & \CPRPL 118 \\
   						      		& \CG 118  & \CG 118  & \CG 118  & \CK & \CPRPL \\

\end{tabular}
\end{table}
\endgroup 

\begingroup
\begin{table}[h]
\setlength{\tabcolsep}{4pt}
\centering
\caption{Total average FLOPs ($\times 10^8$) performed per client per round for different FL tasks, using different optimisers and FL algorithms: \colorbox{red}{FedGBO}, \colorbox{blue}{Mimelite}, \colorbox{green}{MFL} using RMSProp, SGDm and Adam. \colorbox{gray}{AdaptiveFedOpt} and \colorbox{gray}{FedAvg} use SGD. Values shown for $K = 50$ local update steps, to 3 significant figures.}
\begin{tabular}{ c | c c c | c | c }

	Task & RMSProp & SGDm & Adam & SGD & FedProx \\ 
	\hline
	\multirow{3}{*}{CIFAR100} 		& \CR 251 & \CR 251 & \CR 253 & \CK  & \CPRPL \\
						      		& \CB 266 & \CB 266 & \CB 268 & \CK 249 & \CPRPL 251\\
   						      		& \CG 251 & \CG 253 & \CG 255 & \CK & \CPRPL \\
  	\hline
   						      		
	\multirow{3}{*}{Sent140} 		& \CR 0.145 & \CR 0.145 & \CR 0.160 & \CK & \CPRPL \\
						      		& \CB 0.150 & \CB 0.150 & \CB 0.165 & \CK 0.130 & \CPRPL 0.145 \\
   						      		& \CG 0.145 & \CG 0.160 & \CG 0.175 & \CK & \CPRPL \\
  	\hline

   	\multirow{3}{*}{FEMNIST} 		& \CR 161 & \CR 161 & \CR 162 & \CK & \CPRPL \\
						      		& \CB 178 & \CB 178 & \CB 179 & \CK 160 & \CPRPL 161 \\
   						      		& \CG 161 & \CG 162 & \CG 163 & \CK & \CPRPL \\
  	\hline
   				      		
	\multirow{3}{*}{Shakespeare}	& \CR 592 & \CR 592 & \CR 592 & \CK & \CPRPL \\
						      		& \CB 2650 & \CB 2650 & \CB 2650 & \CK 591 & \CPRPL 592 \\
   						      		& \CG 592 & \CG 592 & \CG 592 & \CK & \CPRPL \\

\end{tabular}
\end{table}
\endgroup

\vfill

\end{document}